\documentclass{ecai}
\usepackage{graphicx}
\usepackage{latexsym}

\usepackage{amsmath}
\usepackage{amsthm}

\usepackage{amsfonts}       
\usepackage{nicefrac}       
\usepackage{microtype}      
\usepackage{bbm}
\usepackage{subcaption}
\usepackage{comment}
\theoremstyle{definition}
\usepackage{comment}
\newtheorem{definition}[theorem]{Definition}

\begin{document}

\begin{frontmatter}

\title{Understanding the Spectral Bias of Coordinate Based MLPs Via Training Dynamics}

\author[A]{\fnms{John}~\snm{Lazzari}\thanks{Corresponding Author. Email: jcl19h@fsu.edu.}\orcid{0000-0002-2276-3741}}
\author[B]{\fnms{Xiuwen}~\snm{Liu}}

\address[A]{Florida State University, Department of Mathematics}
\address[B]{Florida State University, Department of Computer Science}

\begin{abstract}

Spectral bias is an important observation of neural network training, stating that the network will learn a low frequency representation of the target function before converging to higher frequency components. This property is interesting due to its link to good generalization in over-parameterized networks. However, in low dimensional settings, a severe spectral bias occurs that obstructs convergence to high frequency components entirely. In order to overcome this limitation, one can encode the inputs using a high frequency sinusoidal encoding. Previous works attempted to explain this phenomenon using Neural Tangent Kernel (NTK) and Fourier analysis. However, NTK does not capture real network dynamics, and Fourier analysis only offers a global perspective on the network properties that induce this bias. In this paper, we provide a novel approach towards understanding spectral bias by directly studying ReLU MLP training dynamics. Specifically, we focus on the connection between the computations of ReLU networks (activation regions), and the speed of gradient descent convergence. We study these dynamics in relation to the spatial information of the signal to understand how they influence spectral bias. We then use this formulation to study the severity of spectral bias in low dimensional settings, and how positional encoding overcomes this. 

\end{abstract}

\end{frontmatter}

\section{Introduction}

In the last several years, neural networks have increasingly been employed to provide a way to learn representations that generalize well in dense, low dimensional domains. Specifically in the field of computer graphics, multi-layer perceptrons (MLPs) with ReLU activations have been vital for applications such as neural radiance fields (NeRF)~\cite{Mildenhall2020nerf} and shape occupancy~\cite{Mescheder2018occupancy}. These networks are referred to as coordinate based MLPs, since they input dense, low dimensional coordinates, and regress the corresponding representation of color, shape, or density for various visual signals. For simplicity, we define a coordinate based MLP as a standard ReLU MLP whose task is regressing any low dimensional signal (e.g. a 1D sinusoid, 2D image, or 3D scene), where inputs \(\boldsymbol{\mathrm{v}} \in \mathbb{R}^d\) (typically \(d\leq3\)) are referred to as coordinates. Coordinate based MLPs are intriguing due to their severe spectral bias, meaning they are practically incapable of learning high frequency target signals~\cite{Mildenhall2020nerf}. This limitation can be overcome through a positional encoding \(\gamma(\boldsymbol{\mathrm{v}}) \in \mathbb{R}^{2dL}\) of coordinates comprised of high frequency sinusoids:

\begin{figure}
  \centering
  \def\twidth{5}
  \subfloat[Coordinates]{%
  \includegraphics[width=0.155\textwidth]{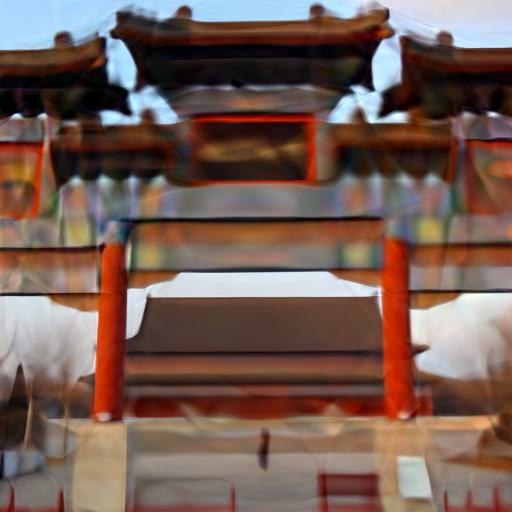}%
  }\hfil
  \subfloat[Encoding L=4]{%
  \includegraphics[width=0.155\textwidth]{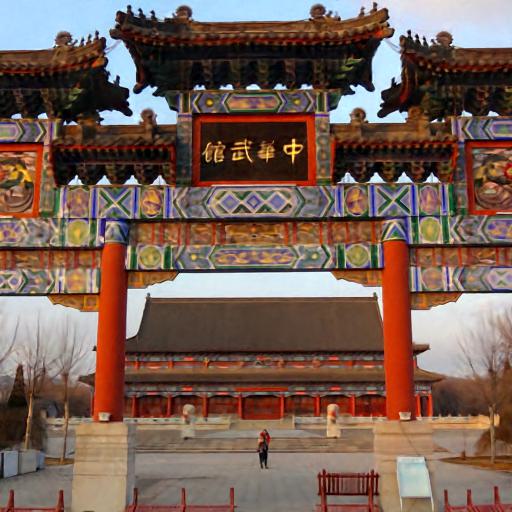}%
  }\hfil
   \subfloat[Encoding L=8]{%
  \includegraphics[width=0.155\textwidth]{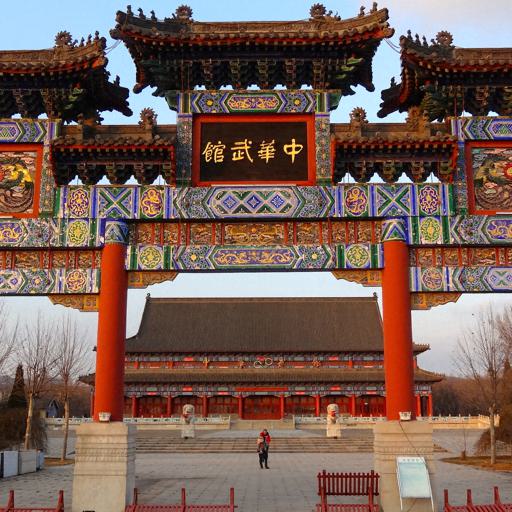}%
  }
  \caption{Visualization of the large spectral bias induced by dense low dimensional 2D coordinates (a), which is overcome by a high frequency positional encoding (b-c). For this task, the coordinate based MLP is tasked with regressing the RGB value corresponding to the 2D pixel location, or its sinusoidal encoding. Coordinate based inputs have difficulty converging to the high frequency components of the image, and can only generate a smooth, low frequency representation.}
  \label{fig:visualization}
\end{figure}

\[\gamma(\boldsymbol{\mathrm{v}}) = \left[\sin(2^0\pi \boldsymbol{\mathrm{v}}),\cos(2^0\pi \boldsymbol{\mathrm{v}}),\cdots,\sin(2^L\pi \boldsymbol{\mathrm{v}}),\cos(2^L\pi \boldsymbol{\mathrm{v}})\right],\]
where \(L \in \mathbb{N}\) determines the maximum frequency as well as encoding dimensionality~\cite{tancik2020fourier,Rahimi2007}. This behavior raises two important questions: What properties of neural network training dynamics induce spectral bias, and why does it become so severe in dense, low dimensional settings?

Spectral bias is the behavior such that neural networks learn a simpler, lower frequency representation of the target function before converging to the high frequency components, or finer details (see Figure 1 for an example). This phenomenon is of interest due to its impact on generalization, since it is believed that an implicit regularization biased towards lower frequency solutions avoids overfitting in over-parameterized networks~\cite{rahaman2019spectral}. While spectral bias is generally believed to aid in generalization, in the coordinate based regime it becomes so severe that the network will essentially underfit the target signal. 

Further understanding the nature of spectral bias will be crucial in determining its impact on generalization. So far, Neural Tangent Kernel (NTK)~\cite{jacot2018neural} and Fourier analysis have been the primary tools for analyzing spectral bias~\cite{tancik2020fourier,rahaman2019spectral}. While NTK models gradient descent dynamics via a kernel method (whose kernel matrix becomes a constant in the limit of very wide layers), the dynamic properties of the real network are not accurately captured. On the other hand, the Fourier decomposition of a ReLU network is bounded by the total number of linear regions as well as its Lipschitz constant~\cite{rahaman2019spectral}, which does provide real network insights. Spectral decay rates of parameter gradients have been found as well. However, this only provides a global perspective on the network properties, mostly through upper bounds, thus the local dynamics of the network that induce spectral bias have yet to be demonstrated. For example, while the total number of linear regions in the network gives a measure of expressive power, it is not necessarily the case that the network will utilize most of these regions, especially if there is a dense sampling of inputs.

In this paper, we take a different approach towards understanding spectral bias. Our contributions are as follows:

\begin{itemize}

    \item We develop a novel framework for understanding spectral bias through real network training dynamics, which incorporates ReLU network computations, gradient descent convergence, as well as the spatial information of the signal. Specifically, we relate the network's expressive capacity (activation regions) with the ability to speed convergence of gradient descent, through a metric termed gradient confusion~\cite{Sankararaman2020}. We find more confusion (slower convergence) when expressive power between inputs is limited, and less confusion (faster convergence) when expressive power is enhanced. This results in slower convergence to the local details of the signal, where inputs are typically restricted to the same or nearby activation regions.

    \item We use this formulation to explore the severity of spectral bias in the coordinate based regime, and how positional encoding overcomes this. We find that positional encoding greatly enhances expressive capacity across the signal, resulting in faster convergence to high frequency components. We provide additional analysis as well, by studying the unique properties of the activation regions as encoding frequency increases, and exposing the dying ReLUs that occur in the low dimensional setting.
    
\end{itemize}

The paper is structured as follows. Section 2 provides related work. In Section 3, we define activation regions and explore how density restricts the expressive capacity of the network. Section 4 discusses gradient confusion, and connects confusion to the network's activation regions. In Section 5, we analyze the properties of the activation regions induced by higher frequency encodings, and Section 6 demonstrates how dense coordinates turn off ReLU neurons during training. Section 7 is a conclusion which details future directions for this approach.

\section{Related Work}

In~\cite{rahaman2019spectral,basri2020frequency}, it was shown that neural networks have a bias towards learning low frequency functions first, referred to as spectral bias, and that a sinusoidal mapping can allow for higher frequencies to be learned faster. This method was adopted by NeRF~\cite{Mildenhall2020nerf} in order to speed converge of high frequency components during novel view synthesis, since a large spectral bias was discovered. As a consequence of this, the relationship between coordinates and positional encoding was analyzed by the same authors using NTK~\cite{tancik2020fourier}. They found that the eigenspectrum of the NTK decays rapidly with dense coordinates, and widens for positional encoding. This allows for faster convergence along the directions of the corresponding eigenfunctions. Our work can best be seen as an extension of~\cite{tancik2020fourier}, however we focus on real network dynamics.

Aside from its severity in low dimensional settings, the nature of spectral bias has been studied from a theoretical perspective.~\cite{rahaman2019spectral} provide the first rigorous exploration of spectral bias using Fourier analysis. In~\cite{Cao2021}, it was shown how the eigenfunctions of the NTK have their own convergence rate given by their corresponding eigenvalue, using inputs with uniform density on \(\mathbb{S}^1\). The authors in~\cite{basri2020frequency} also utilized NTK to give convergence rates for inputs of non-uniform density in \(\mathbb{S}^1\). There has been a line of work referring to spectral bias as the F-principle~\cite{Xu_2020}, in which Fourier analysis was utilized for high dimensional inputs with soft activations. Other works have attempted to compute spectral bias and determine convergence rates as well~\cite{Kiessling_Thor_2022,Basri2019convergence}. Note that previous works have largely focused on gaining insights from MLPs performing regression based tasks. 

Our analysis of spectral bias utilizes activation/linear regions. Exploring the complexity of functions computable by piece-wise linear networks initially explored in~\cite{pascanu2013}. This approach was expanded upon in~\cite{Montufar2014}, where upper and lower bounds for the maximal number of linear regions was given, and shown to be exponential with depth. In~\cite{Raghu2017}, activation regions were defined, and bounds were computed by utilizing input trajectories \(x(t)\) across the regions. More recently,~\cite{Hanin2019complexity,Hanin2019} attempted to compute practical bounds for the number of activation regions, which was found to be independent of depth.~\cite{Zhang2020EmpiricalSO} provides useful experimental quantities for studying the properties of linear regions in a more practical manner. 

In this paper, we relate expressive capacity to the correlation of gradients during training. We specifically focus on gradient confusion~\cite{Sankararaman2020}, which was found to be higher in deeper networks, making them more difficult to train if their width does not increase concurrently. Other works have utilized correlations between gradients in different settings as well~\cite{Balduzzi2017,Fort2019stiffness,yin2018diversity}.

\section{Expressive Power}

\subsection{Experimental Setting}

We begin our study by exploring activation regions, and providing experimental details. For all results in this paper (unless otherwise stated), we train a four layer MLP with 512 hidden neurons on the image regression task. The network learns to regress the RGB values corresponding to the 2D pixel locations \((x_i, y_i)\) using the MSE loss. We average results over five natural images from the div2k dataset~\cite{Agustsson_2017_CVPR_Workshops} (512x512 images). Regressing these images has been a common benchmark in recent works on positional encoding~\cite{tancik2020fourier,fathony2021multiplicative}. Results averaged over five 1D sinusoids are shown in the Appendix where our findings hold. We use the Adam optimizer~\cite{DiederikAdam} with a learning rate of .001, and mini-batches of size 8192. Experiments are conducted on an NVIDIA A5000 GPU.

In addition to the coordinate based regime, we also conduct the same experiments on real world data for both an MLP and a CNN (only Section 4). We use MNIST~\cite{deng2012mnist}, Fashion MNIST~\cite{Xiao2017FashionMNISTAN}, CIFAR10~\cite{Krizhevsky09learningmultiple} and CIFAR100~\cite{Krizhevsky09learningmultiple}. The MLP is the same size, and more details on the CNN architecture (which mainly follows~\cite{Zhang2020EmpiricalSO})  and the experimental setup can be found in the Appendix. We show results for the classification task simply to demonstrate that the behavior is similar to the coordinate based regime, where spectral bias can be easily interpreted. However, spectral bias has been largely unexplored thus far in the classification setting (see Related Work). Because of this, we do not make strong claims on the spectral bias in classification, since a more in-depth analysis is required. Instead, we discuss our results which may be valuable for future works. 

\begin{figure*}[ht]
  \centering
  \def\twidth{0.5}
  \subfloat[Coordinate Activation \\ Regions]{%
  \includegraphics[width=0.22\textwidth]{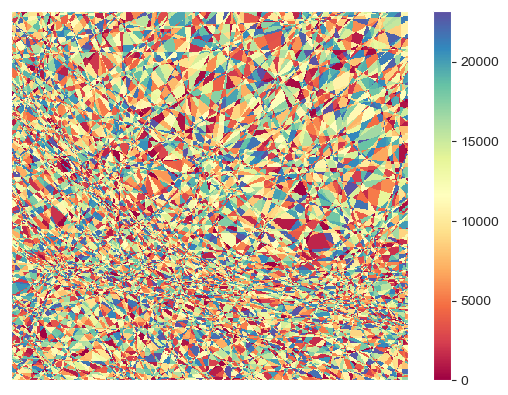}%
  }\hfil
  \subfloat[Encoding Activation \\ Regions]{%
  \includegraphics[width=0.22\textwidth]{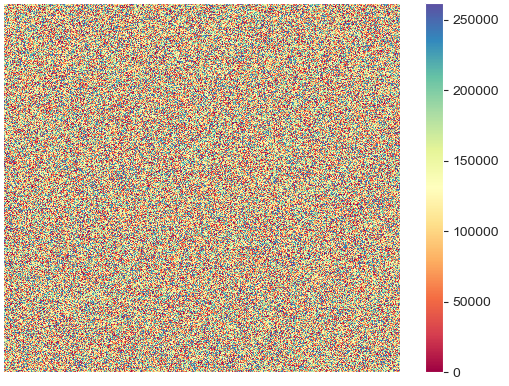}%
  } \hfil
  \subfloat[Training Loss]{%
  \includegraphics[width=0.23\textwidth]{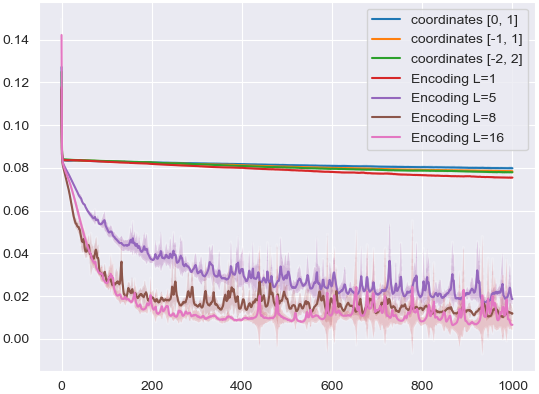}%
  } \hfil
  \subfloat[Activation Region Growth]{%
  \includegraphics[width=0.23\textwidth]{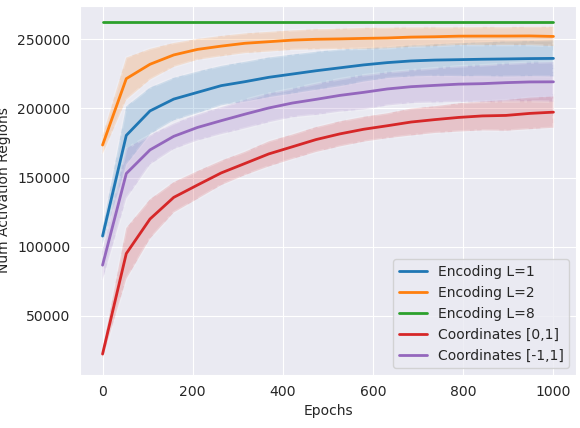}%
  }
  \caption{Visualization of the regions in which each input belongs to (a-b) (colored according to unique region) at the beginning of training. Many coordinates lie in the same activation regions, but positional encoding allows for each input to lie in a unique activation region. In (c), we show the training losses of each method, and in (d) we plot the number of unique activation patterns created as training progresses. We only focus on the activation patterns utilized for the given dataset (512x512 total regions), and demonstrate how coordinate based inputs and low frequency encodings are unable to map each input to a unique activation region throughout training.}
  \label{fig:activation_regions}
\end{figure*}

\begin{figure}
  \centering
  \def\twidth{0.1}
  \subfloat[Coordinates]{%
  \includegraphics[width=0.163\textwidth]{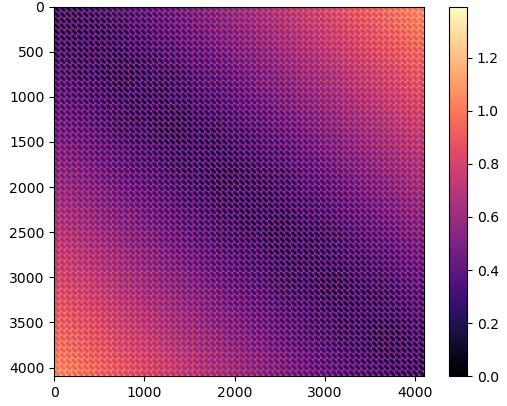}%
  }
  \subfloat[Encoding L=5]{%
  \includegraphics[width=0.163\textwidth]{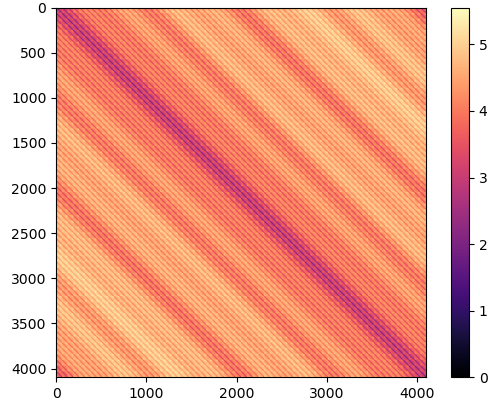}%
  }
  \subfloat[Encoding L=16]{%
  \includegraphics[width=0.163\textwidth]{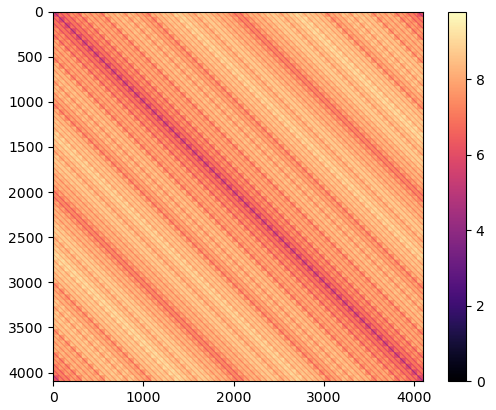}%
  }
  \caption{Matrix of coordinate distances (a) and encoding distances (b-c) for different frequencies. Positional encoding induces a kernel with a strong diagonal since it is stationary. }
  \label{fig:distances}
\end{figure}

\subsection{Activation Regions}

We analyze the expressive power of a neural network with dense and uniform inputs through its activation regions. We begin by defining activation regions and patterns. Let \(f_{\theta} : \mathbb{R}^{n_{\mathrm{in}}} \rightarrow \mathbb{R}^{n_{\mathrm{out}}}\) be a continuous piece-wise linear function given by a ReLU network containing \(L\) hidden layers with \(N\) hidden neurons per layer. The network is defined as the composition of affine transformations \(f_l(x) = \boldsymbol{W}^l x + \boldsymbol{b}^l\) with ReLU activations \(\sigma = \max(0, f_l)\) such that
\[f_{\theta} = f_{\mathrm{out}} \circ \sigma \circ f_L \circ \sigma \circ \ldots \circ \sigma \circ f_1(x),\] 
where \(\theta\) denotes the vector of trainable parameters. Each neuron \(z^l_i(x) = \boldsymbol{W}^l_i x + \boldsymbol{b}^l_i\) composed with \(\sigma\) denotes a hyperplane equation with the scalar determined by the bias. The collection of these hyperplanes in the first layer gives a hyperplane arrangement in \(\mathbb{R}^{n_{\mathrm{in}}}\), splitting the input space into pieces that compute distinct linear functions. As the image of each distinct linear function in the preceding layer \(l-1\) is uniquely partitioned by the following neurons \(z^l_i(x)\), the \(N-1\) dimensional hyperplanes in layers \(l > 1\) will appear to bend. Overall, this leads to a ReLU network partitioning the input space into convex polytopes on which unique linear functions are computed. One way to evaluate these polytopes is through activation regions, which are determined by the network's activation patterns, defined as

\begin{definition}[Activation Pattern] Let \(f_{\theta}\) be a ReLU network, and \(z^l_i(x)\) denote the pre-activation of a neuron in the \(l\)th layer. Then, an activation pattern \(\mathcal{A}\) of the network is a vector in \(\{0, 1\}^{\# \mathrm{neurons}}\) composed of neuron activations such that \(0\) is assigned if \(z^l_i(x) < 0\) and \(1\) if \(z^l_i(x) > 0\).
\end{definition}

\begin{definition}[Activation Region] For a ReLU network \(f_{\theta}\), the activation regions \(\mathcal{R}\) of \(f_{\theta}\) are the sets of input samples that correspond to the same \(\mathcal{A}\),
\begin{multline*}
\mathcal{R}(f, \mathcal{A}) = \{ x \in \mathbb{R}^{n_{in}} \;\; | \;\; \mathbbm{1}_{\mathbb{R}^{+}}(\mathrm{ReLU}(f^l(x))) = a_l, \;\; \\ 
\forall l \in 1,...,k, \;\; \forall a_l \in \mathcal{A} \}
\end{multline*}
where \(a_l\) is the activation pattern of layer \(l\).
\end{definition}

Analyzing activation regions are important because they provide a measure of the expressive power of the network, or the complexity of functions it can compute. In~\cite{Hanin2019}, a realistic upper bound on the maximum number of activation regions created by a ReLU network architecture was given as 
\begin{equation}
\frac{\#\mathrm{Regions \;in \;\mathcal{F} \;intersecting\; \mathcal{C}}}{\mathrm{vol}(\mathcal{C})} \leq \frac{(TN)^{n_{\mathrm{in}}}}{n_{\mathrm{in}}!}
\end{equation}
where \(N\) is the number of neurons, \(T\) is a constant, and \(\mathcal{C}\) is a cube in input space.

Equation (1) is determined by the number of neurons and the input dimension. Therefore, the dimensionality of the inputs themselves show that the network with positional encoding should have more expressive capacity. However, the density of input sampling is a major factor as well, therefore we also analyze how the network utilizes its activation regions across the signal, since focusing solely on the total number of regions would not provide insight on the local dynamics of the network. We can define the set of interest on a dataset \(\mathcal{D} = \{x_i, y_i\}^n_{i=1}\), given in terms of the corresponding activation patterns as \[\mathcal{L_{\mathcal{D}}} = \{ \mathbbm{1}_{\mathbb{R}^{+}}(\mathrm{ReLU}(f^l(x))) \; | \; x \in
\mathcal{D}, \forall l \in 1,...,k \},\] 
equipped with the hamming distance \(\sum_{i=1}^n|a_i - b_i|, \;\) for \(a, b \in \mathcal{L_{\mathcal{D}}}\) to determine their distinctiveness. We later use hamming distance as a proxy for determining expressive power between inputs (higher distance, more expressive power).

\subsection{Comparisons}

In Figure 2, we get a glimpse of the limitations imposed by dense, low dimensional inputs. The number of activation regions utilized with coordinates is lower than the number of elements in the dataset (Fig. 2 (a,b)), meaning many inputs will be regressed using the same linear function. This restriction holds throughout training (Fig. 2(d)). On the other hand, positional encoding can easily map each element to a unique activation region using the same architecture. Overall, we can see that coordinate based inputs have very little expressive power during training in comparison to positional encoding.

The reasoning for this reduced expressive power is a result of both input density and a lack of activation regions in low dimensions. Coordinates impose a grid over a subset of \(\mathbb{R}^d\) in which the resolution, given by their spacing, is very fine-grained. Therefore, fitting enough neurons in the first layer such that a substantial amount of hyperplanes can separate each input becomes difficult. If the input to layers \(l > 1\) are restricted to the images of similar (or the same) linear functions across the dataset, then this severely restricts the overall capacity of the network even as the depth increases (see Appendix). With positional encoding, the density is alleviated since the distance \(\lambda = ||\gamma(x_i) - \gamma(x_j)||\) for each \(x_i, x_j\) is scaled by the frequency, which simultaneously raises its dimensionality and generates more activation regions. The density of inputs can be interpreted through the induced kernel function, which is known to be stationary for positional encoding~\cite{tancik2020fourier}. This means that the similarity between inputs is only dependent on their distance, inducing a strong diagonal that is missing with coordinate based inputs (Figure 3). 

For intuition, we can also relate the density of inputs to the frequency components of the target function. Higher density results in a target function containing higher frequency components, and vice versa. Thus, positional encoding can also be viewed as generating a lower frequency function, which better disperses inputs across available activation regions. This is not possible with the fine-grained sampling of coordinates, as shown in Figure 2.

\section{High Gradient Confusion}

In~\cite{Sankararaman2020}, the effect of network architecture on speed of convergence was modeled using gradient confusion, a measure determining the correlation of gradients for differing inputs. It was shown that lower confusion during training can speed convergence of SGD, or make training the network easier. Confusion occurs when two objective functions \(\mathcal{L}(f_{\theta}(x_i), y_i)\) and  \(\mathcal{L}(f_{\theta}(x_j), y_j)\), \(i \neq j\), have gradients such that \(\langle \nabla  \mathcal{L}(f_{\theta}(x_i), y_i), \nabla \mathcal{L}(f_{\theta}(x_j), y_j) \rangle < 0\). This creates a disagreement on the direction the parameters need to move, slowing down convergence. Convergence rates of SGD were given through the confusion bound \(\eta \geq 0\), given as 
\[\langle \nabla  \mathcal{L}(f_{\theta}(x_i), y_i), \nabla \mathcal{L}(f_{\theta}(x_j), y_j) \rangle \geq -\eta\]
for all \(i \neq j\) and fixed \(\theta\). We focus on this metric since we are interested in the speed at which gradient descent converges and its relation to the spatial information of the signal.

\subsection{Intuition for Higher Confusion}

In this section, we aim to build intuition as to why limited expressive power will induce higher amounts of confusion when the target values oscillate rapidly. Assume we are regressing a 1D signal, and let \(f^{\mathcal{A}}\) be a linear function given by a ReLU network corresponding to an activation pattern \(\mathcal{A}\). Assume two dense inputs \(x_i, x_j \in [0, 1]\) belong to the same \(\mathcal{R}(f, \mathcal{A})\). We can then write \(x_j = c_0x_i\), and \(\mathrm{sgn}(w_i^{(l)T}(c^{l-1}_i \odot x^{l-1}_i) + b_i^l) = \mathrm{sgn}(w_i^{(l)T}x^{l-1}_i + b_i^l)\) must hold for all neurons, where \(c^l\) is a vector of positive values. Therefore,

\begin{equation}
    \begin{split}
    \left<\nabla_{\theta}f^{\mathcal{A}}(x_i), \nabla_{\theta}f^{\mathcal{A}}(x_j)\right> &= \\
    &\left<\nabla_{\theta}f^{\mathcal{A}}(x_i), d \odot \nabla_{\theta}f^{\mathcal{A}}(x_i)\right>  > 0,
    \end{split}
\end{equation}
where \(d \in \mathbb{R}^{\#\mathrm{weights}}\) are the positive values that scale the hidden outputs.

We will now evaluate using Mean Squared Error (MSE) loss \[\frac{1}{N}\sum_{i=1}^N (y_i - f(x_i))^2.\]
We have the gradient of MSE with respect to the parameters for a single input \(x\) as  
\[\nabla_{\theta} \mathcal{L}(f^{\mathcal{A}}(x), y) = -2(y - f^{\mathcal{A}}(x))\nabla_{\theta}f^{\mathcal{A}}(x).\] 
From (2), we can see that confusion will be directly induced by residual \(y - f^{\mathcal{A}}(x)\). If \(y_i\) is highly oscillatory for each \(x_i \in \mathcal{R}(f, \mathcal{A})\), then confusion will occur in any situation where \(y_i > f^{\mathcal{A}}(x_i)\) and \(y_j < f^{\mathcal{A}}(x_j)\). Since the direction of the gradient update for each weight is essentially determined by the residual, confusion can only be reduced if the target values change linearly or possibly if constant.

\begin{figure*}[ht]
    \centering
    \includegraphics[width=.99\textwidth]{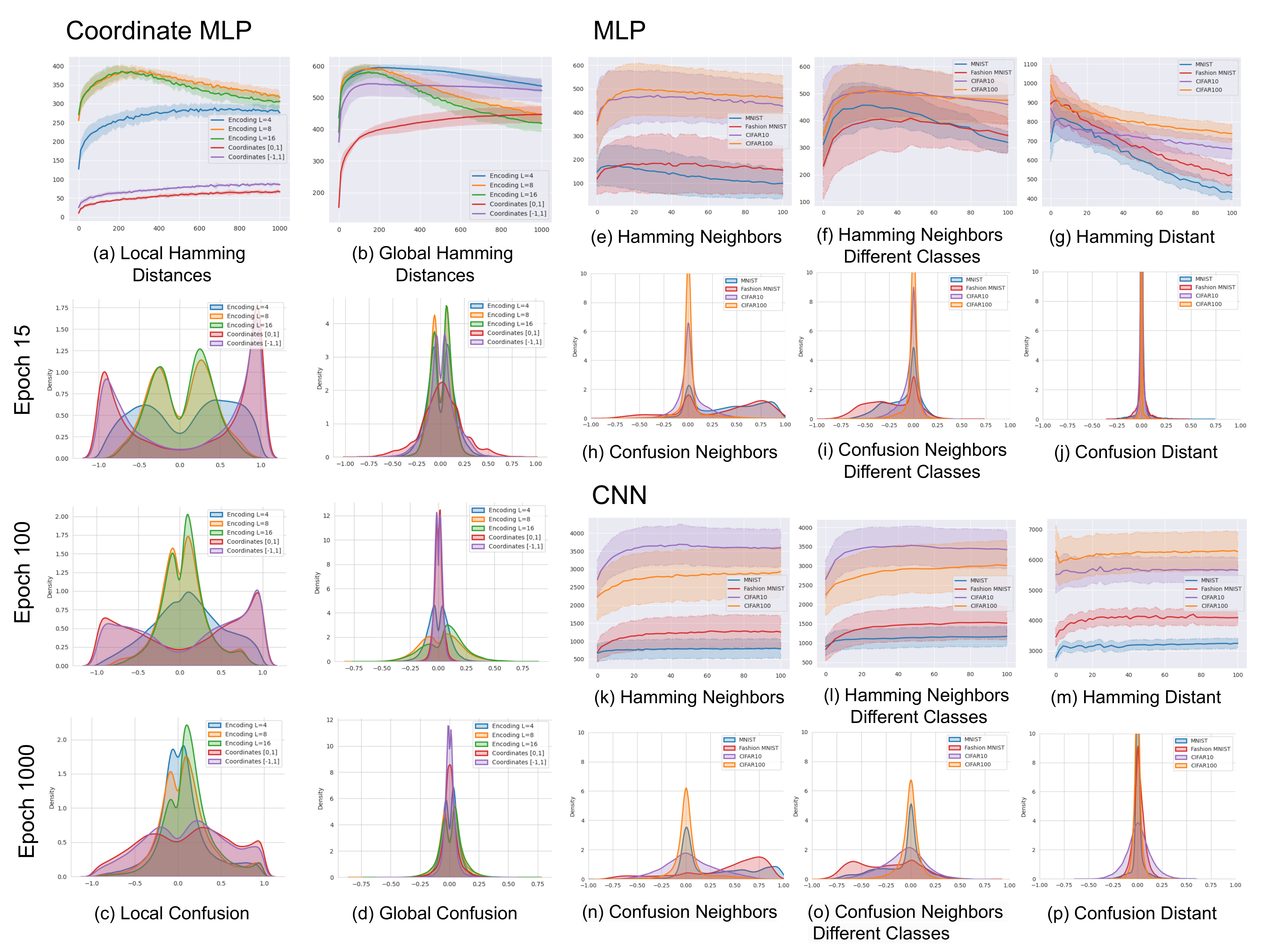}
    \caption{Hamming distances and confusion densities in the coordinate based regime as well as the classification setting. For the coordinate based regime, we sample inputs in 100 different 50x50 local regions of the image, as well as 50k distant inputs. Confusion and Hamming distances are computed between pairs in their respective samples. This amounts to 50k pairs locally and globally. The same results are conducted for an MLP and CNN on various classification datasets, where the local pairs are the \(k=25\) nearest neighbors of a randomly sampled input (which may likely contain mostly the same labels for certain datasets), as well as the \(k=25\) nearest neighbors that are strictly of a different class. The distant pairs are the \(k=25\) furthest inputs to a given sample. Each experiment uses a total 62.5k pairs sampled throughout the dataset. We show 1000 epochs for coordinates, and 100 for classification.}
    \label{fig:narrowness}
\end{figure*}

Now, assume we have inputs \(x_i, x_j \in [0,1]\) such that they are regressed by two separate linear function \(f^{\mathcal{A}}\) and \(f^{\mathcal{B}}\) of patterns \(\mathcal{A}\) and \(\mathcal{B}\). Let \(S = \{ w^l_i \; | \; \mathcal{A}^l_i, \mathcal{B}^l_i = 1\}\) denote the set of commonly activated weights between layers. The inner product of the gradients will only depend on the parameters in \(S\) since all other weights will cancel out due to ReLU \[\langle \nabla_{S}\mathcal{L}(f^{\mathcal{A}}(x_i), y_i), \nabla_{S}\mathcal{L}(f^{\mathcal{B}}(x_j), y_j)\rangle.\] While the residual of \(\nabla_{S}\mathcal{L}\) may still be highly oscillatory, it is not necessarily the case that \(\langle \nabla_{S}f^{\mathcal{A}}(x_i), \nabla_{S}f^{\mathcal{B}}(x_j) \rangle > 0\), as backpropogation will utilize weights \(w^l_i \not\in S\), the number of which is influenced by the Hamming distance between activation patterns. We can view this as the network utilizing information from distinct activation regions when optimizing shared weights. Essentially, the network can configure the activation regions such that updating shared weights in the same direction can be beneficial for minimizing the loss. In this case, the impact of the residual in determining the gradient directions is largely reduced, while the directions for inputs restricted to a single region (or nearby regions) will be highly influenced by the target values.

\subsection{Spectral Bias}

The relationship between expressive capacity and confusion is demonstrated by analyzing the confusion densities and Hamming distances in Figure 4. The Hamming distance is a measure of the expressive power between inputs, as it quantifies the dissimilarity of their activation patterns. We compute both confusion and Hamming distance for inputs sampled within local regions of input space and globally (spaced out intervals) across input space, which should generally correspond to the high and low frequency components respectively (in the coordinate based regime). The pair-wise gradient correlations between inputs at different stages of training are then calculated using cosine similarity, and their densities are plotted. A wider density or higher concentrations in the negative range indicate more confusion, while a density more centered around zero or positive values suggests less confusion. This corresponds to orthogonal or positively correlated gradient directions between inputs.

 In Figure 4, we display three key aspects of our observations. More experimental details are explained in the caption of Figure 4. We first focus on the coordinate based regime as it directly relates to spectral bias:
 
 \begin{enumerate}
     \item Both positional encoding and coordinates are less effective in representing high-frequency components than low-frequency components, as measured by the lower Hamming distances between activation patterns for locally sampled inputs (Fig. 4(a)) in comparison to globally sampled inputs (Fig. 4(b)). As a result, during training, the model gets confused more often when dealing with high-frequency components than low-frequency components (Fig. 4(c)-(d)).
     \item Positional encoding endures the least amount of confusion for the high frequency components due to its enhanced expressive capabilities (Fig. 4(a)), which suggests faster convergence.
     \item Coordinates generate a large disparity in the quantity of confusion between input samples, which does not occur with positional encoding.
 \end{enumerate}

\begin{figure}
    \centering
  \subfloat[L=4 \\(epoch 1)]{\includegraphics[width=0.15\textwidth]{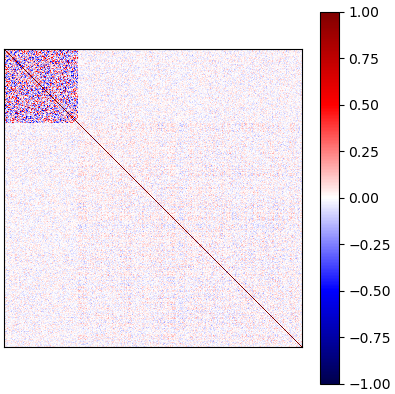}} \hfil
  \subfloat[L=8 \\(epoch 1)]{\includegraphics[width=0.15\textwidth]{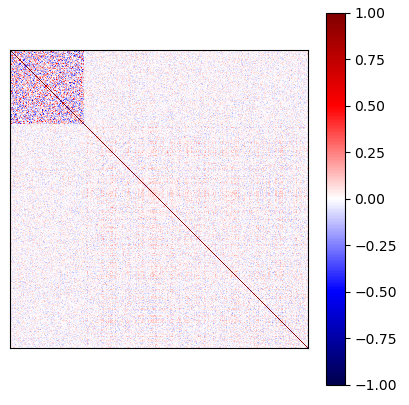}}
  \hfil
  \subfloat[L=16 \\(epoch 1)]{\includegraphics[width=0.15\textwidth]{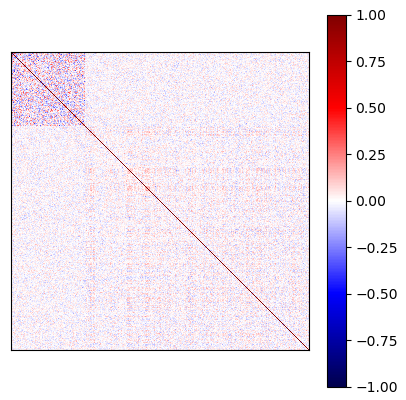}}
  \vspace{.01mm}
  \subfloat[L=4 \\(epoch 2500)]{\includegraphics[width=0.15\textwidth]{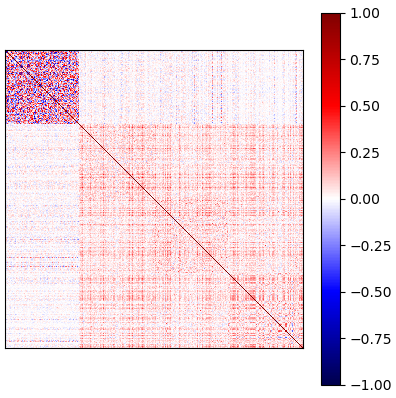}} \hfil
  \subfloat[L=8 \\(epoch 2500)]{\includegraphics[width=0.15\textwidth]{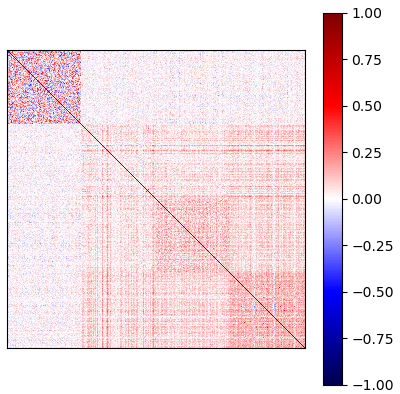}} \hfil
  \subfloat[L=16 \\(epoch 2500)]{\includegraphics[width=0.15\textwidth]{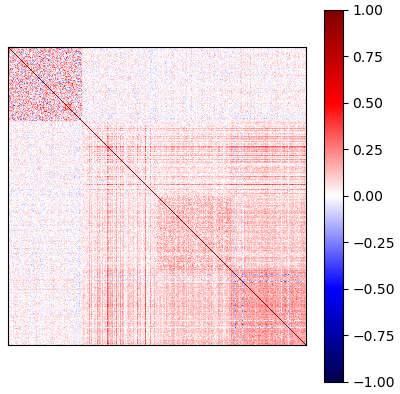}}
  \caption{Cosine similarity between hyperplane normals during training across layers. The normal to each hyperplane is simply given as the 
  corresponding weights (Conducted on a single image, holds for other images).}
\end{figure}

Overall, these results demonstrate how each input will likely have its convergence obstructed by its nearest neighbors, caused by the similarity in their activation patterns which induces higher concentrations of confusion. In contrast, distant inputs sampled globally across the signal suffer less confusion since more expressive power can be utilized, resulting in harmonious gradient updates. Thus, the network converges faster to the overall structure of the target function than to the local details, resulting in the observed low-frequency representations. Positional encoding allows the network to utilize its enhanced expressive power in an effective manner, granting quicker convergence to the high frequency components. Moreover, the significant disparity in confusion when using coordinates (between varying frequency components) can be attributed to a larger spectral bias.

For the classification tasks, a similar behavior occurs for both architectures. The nearest neighbors induce more confusion than the distant inputs, with neighbors of strictly separate classes inducing the most confusion. The Hamming distances between activation patterns follow accordingly. We reiterate that, while our method provides insight on the spectral bias when regressing signals (slower convergence to rapidly changing local details), directly relating these results to the classification setting should require a deeper analysis. So far, the authors of~\cite{rahaman2019spectral} have utilized a binarized version of the MNIST dataset (only two labels) with increasing label noise to demonstrate a spectral bias, but this setting is limited compared to multi-class classification, and the objective was MSE. In~\cite{Xu_2020}, spectral bias was analyzed in the classification setting using what they defined as response frequency, given through the non-uniform discrete Fourier transform (NUDFT) on a dataset \(\{(x_i, y_i)^N_{i=0}\}\) of images and labels. In this case, we can potentially relate the high frequency components of the dataset to the nearest neighbors of different classes, where confusion is concentrated. Further comparing our results to the methods utilized in previous works may provide insight on the properties that induce spectral bias in the high dimensional classification setting. 

\section{Encoding Frequency and Activation Region Properties}

We now explore the distinct activation region dynamics induced by higher frequency positional encodings. Doing so can give more insight into the distinct solutions the network discovers, and we utilize experiments inspired by previous works ~\cite{Hanin2019complexity,Zhang2020EmpiricalSO,Novak2018SensitivityAG,Neyshabur2017generalization}. We start by analyzing the correlation between hyperplane directions, given as \(\nabla_x z(x)\), by taking the cosine similarity between all weights (Figure 5). We find that higher frequency encodings allow for an increase in orthogonal hyperplane directions in the beginning of training (higher dimensional), then become increasingly parallel as training progresses. Specifically, Figure 5 demonstrates that hyperplanes are less negatively correlated in the first layer as the encoding frequency increases, which affects the following layers accordingly.

\begin{figure}[h]
    \centering
    \includegraphics[width=.49\textwidth]{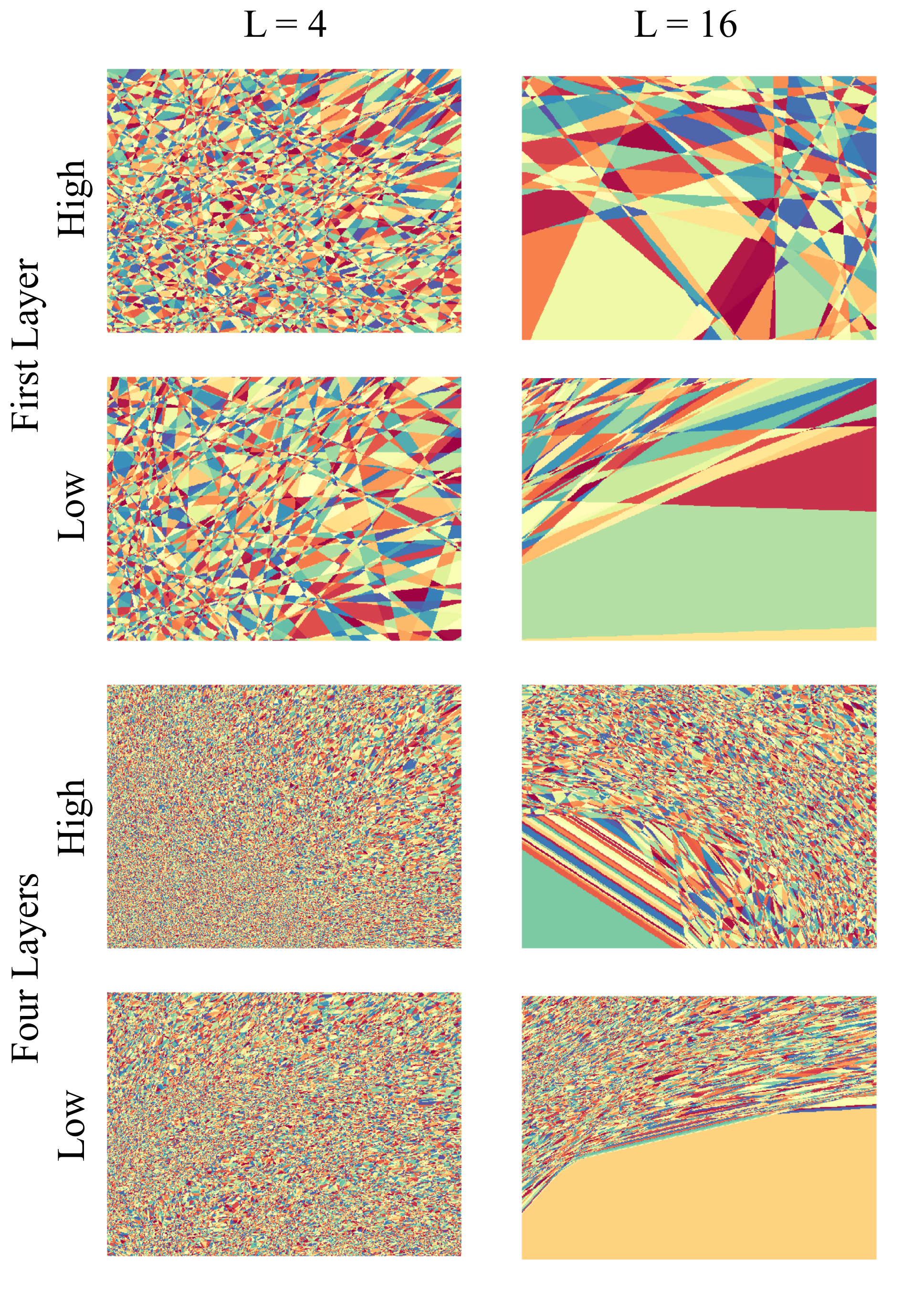}
    \caption{2D slice of activation regions for high and low frequency encoding after 2500 epochs. Images were found by moving along a 2D plane (all other dimensions fixed) at the origin, with "Low' denoting a 2D plane along the dimensions of the low frequency features of the encoding, and "High" the high frequency features. Visualization done on single image.}
    \label{fig:slice}
\end{figure}


We additionally find that activation regions will expand (or increase in volume) with encoding frequency later in training, which may cause a reduction in the distinctiveness between activation patterns across the dataset. This reduction is shown in Figure 4 for higher frequency encodings (L=8 and L=16), as there is a large dip in the mean Hamming distance. Notice that the low frequency encoding (L=4) begins to contract the Hamming distances as well, but at a much slower rate. To further demonstrate that an increase in encoding frequency results in wider activation regions, we visualize 2D slices of the activation regions for a trained network in Figure 6, where the distinction is shown. 

Lastly, in Figure 7 we plot the distance of each input to the nearest boundary, given as
\[d(x, \mathcal{B}_f) = \min_{z(x) \in f}|z(x) - b_z| / ||\nabla z(x)||,\] 
which was originally shown in~\cite{Hanin2019complexity}. This distance measures the sensitivity of neurons to the given input samples, or in other words their sensitivity to boundary transitions. Note that this distance may be influenced by the size of the regions, but does not provide a direct measure. Rather, it is possible that areas of input space corresponding to wider activation regions would suggest a low transition density in this vicinity (relative to architecture and input size), which is a separate sensitivity metric~\cite{Novak2018SensitivityAG}.

\begin{figure}
    \centering
    \includegraphics[width=.32\textwidth]{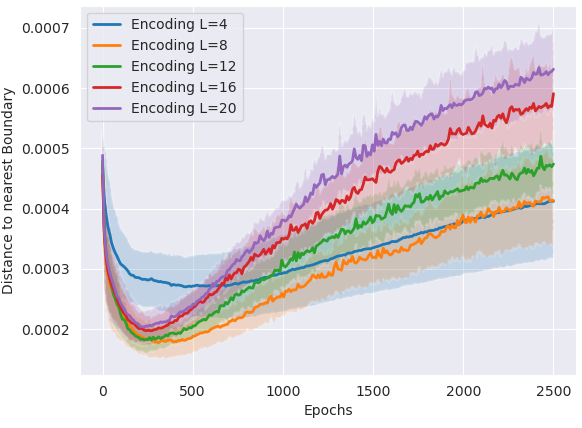}
    \caption{The mean distance of training points to boundaries \(\mathcal{B}_f\) of ReLU neurons. This gives a measure of the sensitivity of neurons to the inputs samples during training, and how this fluctuates.}
    \label{fig:narrowness}
\end{figure}

In Figure 7, the higher frequency encodings first have a larger contraction of this distance, then rapidly increase it during training, indicating a shift from high to low sensitivity. This measure is interesting as it was used in~\cite{Hanin2019complexity} for the MNIST dataset, where a similar behavior was found. They discovered that the network will quickly contract the distances, which is correlated with better generalization performance towards the end of training. In this setting, we also notice an increase after this contraction which is larger for higher frequency encodings. 

Overall, networks using higher frequency encodings tend to find solutions in which the hyperplanes are highly correlated, the volume of the activation regions increase, and the sensitivity of neurons to the input samples is reduced later in training. These properties may induce faster convergence to the target signal as shown in Section 4, which lower frequency encodings cannot accomplish. However, they may also be linked to worse generalization performance, since higher frequency encodings are known to overfit~\cite{tancik2020fourier}. While connecting each of these distinct properties is difficult due to the complex dynamics of the regions, doing so may provide insight towards the generalization of positional encoding in future works.

\section{Dying ReLU}

Although coordinate based networks slowly increase the number of activation regions throughout training as shown in Figure 2 (d), there is a surprising alternative that is simultaneously occurring. That is, the network is increasingly \emph{shutting off} ReLU neurons during training, limiting its full expressive potential while simultaneously attempting to utilize different linear functions with available neurons. We believe this is due to the exponential increase in parameter norms that occur when training on high frequency target functions, which can cause issues due to the density of the data. Note this only occurs with coordinates, and not with positional encoding.

We computed the parameter norms \(\prod_{n=1}^L ||\boldsymbol{W}^{(k)}||\) across layers, where \(|| \cdot ||\) is the spectral norm found by the maximum singular value, displayed in Figure 8. This norm represents the upper bound of the Lipschitz constant \(L_f\) of the network, and it may be slightly counter-intuitive to see that this bound is higher for the network with a larger spectral bias. In this setting, it may be the case that larger parameters can easily map coordinates to one side of the bias threshold, causing it to be inactive for the rest of training. We can see some intuition for this idea from Figure 8 (d). The number of dead ReLU neurons decreases as the normalizing interval increases, meaning once the sampling becomes less dense there are more active neurons. While this may be the case, there is still the overall trend of an increasing amount of dead neurons, which displays another limitation of ReLU networks in this dense, low dimensional setting. Additionally, this seems to be an issue at initialization, which can perhaps be reduced with proper weight initialization, or other potential measures.

\begin{figure}
  \centering
  \subfloat[Coordinates]{\includegraphics[width=0.225\textwidth]{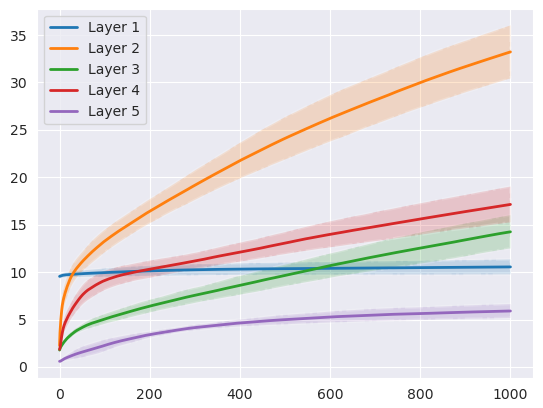}} \hfil
  \subfloat[Positional Encoding \\ L=8]{\includegraphics[width=0.225\textwidth]{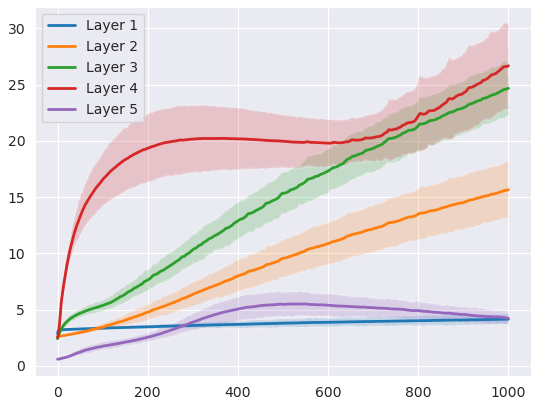}} \hfil \\
  \subfloat[All Methods]{\includegraphics[width=0.24\textwidth]{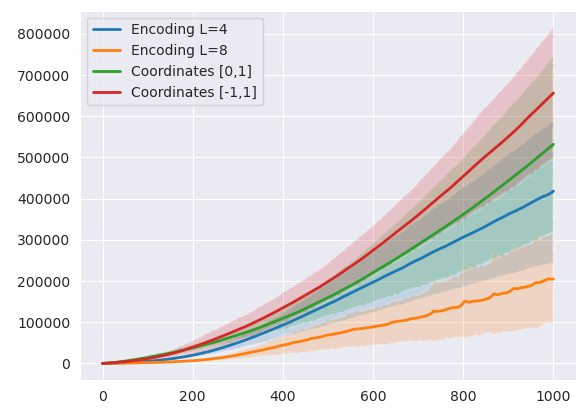}} \hfil
  \subfloat[Dead ReLU Neurons]{\includegraphics[width=0.225\textwidth]{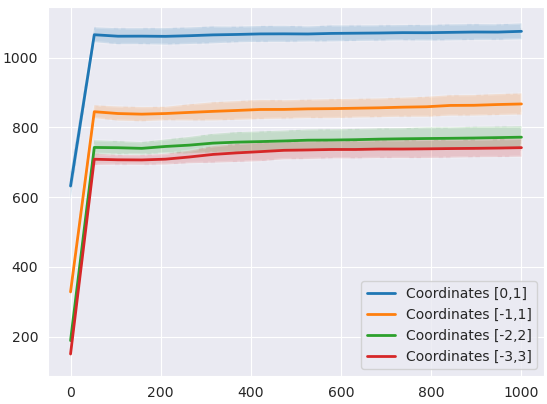}}
  \caption{Spectral norm for both positional encoding (L=8) (b) and coordinates ([0,1]) (a) throughout training (a-c), along with the amount of dead ReLU neurons during training for coordinates (d). There are no dead ReLU neurons for positional encoding.}
  \label{fig:dead_relu}
\end{figure}

\section{Conclusion}

In this paper, we provided a thorough study of spectral bias in the coordinate based setting by using a direct model of training dynamics. This analysis provides the first look into the properties of the network that induce spectral bias without the use of NTK or Fourier analysis. We found that convergence of gradient descent in local regions of input space, which utilize less expressive capacity for neighboring inputs, is slower in comparison to inputs sampled globally across the signal. This results in the high frequency components of the signal being learned after a low frequency representation is achieved, with the severity of spectral bias depending on how the network assigns inputs to activation regions. Positional encoding is able to enhance expressive capacity overall, which can result in faster convergence to the high frequency components. In addition to this, we explored the properties of the activation regions as encoding frequency increases, which gains insight into the differing solutions the network arrives at. Another limitation of ReLU in this low dimensional setting comes from the dying ReLUs, which are most likely caused by the density of inputs and the exponential increase in parameter norms. Lastly, we show that the same general behavior exists in the classification setting with different architectures, which may provide insight for future works which attempt to formally analyze and measure spectral bias in more complex settings. It is also possible that a method such as ours can overcome the computational burdens of previous methods in these complex settings as well, such as Fourier analysis in high dimensional settings.

\bibliography{ecai}

\begin{thebibliography}{10}

\bibitem{kn:Adams85}
L.~Adams, `{m-Step} preconditioned {Gradient} methods', {\em SIAM Journal of
  Scientific and Statistical Computing}, {\bf 6},  452--463, (1985).

\bibitem{kn:Atkin}
P.~Atkin.
\newblock Performance maximisation.
\newblock INMOS Technical Note 17.

\bibitem{kn:daCunha92b}
R.D. {da Cunha} and T.R. Hopkins, {\em The Parallel Solution of Partial
  Differential Equations on Transputer Networks},  96--109, Transputing for
  Numerical and Neural Network Applications, IOS Press, Amsterdam, 1992.
\newblock Also as Report No. 17/92, Computing Laboratory, University of Kent at
  Canterbury, U.K.

\bibitem{kn:daCunha92a}
R.D. {da Cunha} and T.R. Hopkins, {\em The Parallel Solution of Systems of
  Linear Equations using Iterative Methods on Transputer Networks},  1--13,
  Transputing for Numerical and Neural Network Applications, IOS Press,
  Amsterdam, 1992.
\newblock Also as Report No. 16/92, Computing Laboratory, University of Kent at
  Canterbury, U.K.

\bibitem{kn:deCarlini91}
U.~{de Carlini} and U.~Villano, {\em Transputers and parallel architectures --
  message-passing distributed systems}, Ellis Horwood, Chichester, 1991.

\bibitem{kn:Eisenstat81}
S.C. Eisenstat, `Efficient implementation of a class of preconditioned
  {Conjugate Gradient} methods', {\em SIAM Journal of Scientific and
  Statistical Computing}, {\bf 2},  1--4, (1981).

\bibitem{kn:Golub89}
G.H. Golub and C.F. {Van Loan}, {\em Matrix Computations}, Johns Hopkins
  University Press, Baltimore, 2nd edn., 1989.

\bibitem{kn:Johnson83}
O.G. Johnson, C.A. Micchelli, and G.~Paul, `Polynomial preconditioners for
  {Conjugate Gradient} calculations', {\em SIAM Journal of Numerical Analysis},
  {\bf 20},  362--376, (1983).

\bibitem{kn:Modi88}
J.J. Modi, {\em Parallel Algorithms and Matrix Computation}, Oxford University
  Press, Oxford, 1988.

\bibitem{kn:Saad85}
Y.~Saad, `Practical use of polynomial preconditionings for the {Conjugate
  Gradient} method', {\em SIAM Journal of Scientific and Statistical
  Computing}, {\bf 6},  865--881, (1985).

\bibitem{kn:Schofield89}
C.F. Schofield, {\em Optimising {FORTRAN} programs}, Ellis Horwood Publishing,
  Chichester, 1989.

\bibitem{kn:Smith85}
G.D. Smith, {\em Numerical Solution of Partial Differential Equations: Finite
  Difference Methods}, Oxford University Press, Oxford, 3rd edn., 1985.

\end{thebibliography}


\begin{thebibliography}{10}

\bibitem{Agustsson_2017_CVPR_Workshops}
Eirikur Agustsson and Radu Timofte, `Ntire 2017 challenge on single image
  super-resolution: Dataset and study', in {\em The IEEE Conference on Computer
  Vision and Pattern Recognition (CVPR) Workshops}, (July 2017).

\bibitem{Balduzzi2017}
David Balduzzi, Marcus Frean, Lennox Leary, JP~Lewis, Kurt Wan-Duo Ma, and
  Brian McWilliams, `The shattered gradients problem: If resnets are the
  answer, then what is the question?', (2017).

\bibitem{basri2020frequency}
Ronen Basri, Meirav Galun, Amnon Geifman, David Jacobs, Yoni Kasten, and Shira
  Kritchman, `Frequency bias in neural networks for input of non-uniform
  density', in {\em International Conference on Machine Learning}, pp.
  685--694. PMLR, (2020).

\bibitem{Cao2021}
Yuan Cao, Zhiying Fang, Yue Wu, Ding-Xuan Zhou, and Quanquan Gu, `Towards
  understanding the spectral bias of deep learning', in {\em Proceedings of the
  Thirtieth International Joint Conference on Artificial Intelligence,
  {IJCAI-21}}, ed., Zhi-Hua Zhou, pp. 2205--2211. International Joint
  Conferences on Artificial Intelligence Organization, (8 2021).
\newblock Main Track.

\bibitem{deng2012mnist}
Li~Deng, `The mnist database of handwritten digit images for machine learning
  research', {\em IEEE Signal Processing Magazine}, {\bf 29}(6),  141--142,
  (2012).

\bibitem{fathony2021multiplicative}
Rizal Fathony, Anit~Kumar Sahu, Devin Willmott, and J~Zico Kolter,
  `Multiplicative filter networks', in {\em International Conference on
  Learning Representations}, (2021).

\bibitem{Fort2019stiffness}
Stanislav Fort, Paweł~Krzysztof Nowak, Stanislaw Jastrzebski, and Srini
  Narayanan.
\newblock Stiffness: A new perspective on generalization in neural networks,
  2019.

\bibitem{Hanin2019complexity}
Boris Hanin and David Rolnick, `Complexity of linear regions in deep networks',
  in {\em Proceedings of the 36th International Conference on Machine
  Learning}, eds., Kamalika Chaudhuri and Ruslan Salakhutdinov, volume~97 of
  {\em Proceedings of Machine Learning Research}, pp. 2596--2604. PMLR, (09--15
  Jun 2019).

\bibitem{Hanin2019}
Boris Hanin and David Rolnick, {\em Deep ReLU Networks Have Surprisingly Few
  Activation Patterns}, Curran Associates Inc., Red Hook, NY, USA, 2019.

\bibitem{jacot2018neural}
Arthur Jacot, Franck Gabriel, and Cl{\'e}ment Hongler, `Neural tangent kernel:
  Convergence and generalization in neural networks', {\em Advances in neural
  information processing systems}, {\bf 31}, (2018).

\bibitem{Kiessling_Thor_2022}
Jonas Kiessling and Filip Thor, `A computable definition of the spectral bias',
  {\bf 36},  7168--7175, (Jun. 2022).

\bibitem{DiederikAdam}
Diederik~P. Kingma and Jimmy Ba, `Adam: {A} method for stochastic
  optimization', in {\em 3rd International Conference on Learning
  Representations, {ICLR} 2015, San Diego, CA, USA, May 7-9, 2015, Conference
  Track Proceedings}, eds., Yoshua Bengio and Yann LeCun, (2015).

\bibitem{Krizhevsky09learningmultiple}
Alex Krizhevsky, `Learning multiple layers of features from tiny images',
  Technical report, (2009).

\bibitem{Mescheder2018occupancy}
Lars Mescheder, Michael Oechsle, Michael Niemeyer, Sebastian Nowozin, and
  Andreas Geiger.
\newblock Occupancy networks: Learning 3d reconstruction in function space,
  2018.

\bibitem{Mildenhall2020nerf}
Ben Mildenhall, Pratul~P. Srinivasan, Matthew Tancik, Jonathan~T. Barron, Ravi
  Ramamoorthi, and Ren Ng.
\newblock Nerf: Representing scenes as neural radiance fields for view
  synthesis, 2020.

\bibitem{Montufar2014}
Guido~F Montufar, Razvan Pascanu, Kyunghyun Cho, and Yoshua Bengio, `On the
  number of linear regions of deep neural networks', in {\em Advances in Neural
  Information Processing Systems}, eds., Z.~Ghahramani, M.~Welling, C.~Cortes,
  N.~Lawrence, and K.Q. Weinberger, volume~27. Curran Associates, Inc., (2014).

\bibitem{Neyshabur2017generalization}
Behnam Neyshabur, Srinadh Bhojanapalli, David Mcallester, and Nati Srebro,
  `Exploring generalization in deep learning', in {\em Advances in Neural
  Information Processing Systems}, eds., I.~Guyon, U.~Von Luxburg, S.~Bengio,
  H.~Wallach, R.~Fergus, S.~Vishwanathan, and R.~Garnett, volume~30. Curran
  Associates, Inc., (2017).

\bibitem{Novak2018SensitivityAG}
Roman Novak, Yasaman Bahri, Daniel~A. Abolafia, Jeffrey Pennington, and
  Jascha~Narain Sohl-Dickstein, `Sensitivity and generalization in neural
  networks: an empirical study', {\em ArXiv}, {\bf abs/1802.08760}, (2018).

\bibitem{pascanu2013}
Razvan Pascanu, Guido Montufar, and Yoshua Bengio.
\newblock On the number of response regions of deep feed forward networks with
  piece-wise linear activations, 2013.

\bibitem{Raghu2017}
Maithra Raghu, Ben Poole, Jon Kleinberg, Surya Ganguli, and Jascha~Sohl
  Dickstein, `On the expressive power of deep neural networks', in {\em
  Proceedings of the 34th International Conference on Machine Learning - Volume
  70}, ICML'17, p. 2847–2854. JMLR.org, (2017).

\bibitem{rahaman2019spectral}
Nasim Rahaman, Aristide Baratin, Devansh Arpit, Felix Draxler, Min Lin, Fred
  Hamprecht, Yoshua Bengio, and Aaron Courville, `On the spectral bias of
  neural networks', in {\em International Conference on Machine Learning}, pp.
  5301--5310. PMLR, (2019).

\bibitem{Rahimi2007}
Ali Rahimi and Benjamin Recht, `Random features for large-scale kernel
  machines', in {\em Advances in Neural Information Processing Systems}, eds.,
  J.~Platt, D.~Koller, Y.~Singer, and S.~Roweis, volume~20. Curran Associates,
  Inc., (2007).

\bibitem{Basri2019convergence}
Basri Ronen, David Jacobs, Yoni Kasten, and Shira Kritchman, `The convergence
  rate of neural networks for learned functions of different frequencies', in
  {\em Advances in Neural Information Processing Systems}, eds., H.~Wallach,
  H.~Larochelle, A.~Beygelzimer, F.~d\textquotesingle Alch\'{e}-Buc, E.~Fox,
  and R.~Garnett, volume~32. Curran Associates, Inc., (2019).

\bibitem{Sankararaman2020}
Karthik~A. Sankararaman, Soham De, Zheng Xu, W.~Ronny Huang, and Tom Goldstein,
  `The impact of neural network overparameterization on gradient confusion and
  stochastic gradient descent', in {\em Proceedings of the 37th International
  Conference on Machine Learning}, ICML'20. JMLR.org, (2020).

\bibitem{tancik2020fourier}
Matthew Tancik, Pratul Srinivasan, Ben Mildenhall, Sara Fridovich-Keil, Nithin
  Raghavan, Utkarsh Singhal, Ravi Ramamoorthi, Jonathan Barron, and Ren Ng,
  `Fourier features let networks learn high frequency functions in low
  dimensional domains', {\em Advances in Neural Information Processing
  Systems}, {\bf 33},  7537--7547, (2020).

\bibitem{Xiao2017FashionMNISTAN}
Han Xiao, Kashif Rasul, and Roland Vollgraf, `Fashion-mnist: a novel image
  dataset for benchmarking machine learning algorithms', {\em ArXiv}, {\bf
  abs/1708.07747}, (2017).

\bibitem{Xu_2020}
Zhi-Qin~John Xu, `Frequency principle: Fourier analysis sheds light on deep
  neural networks', {\em Communications in Computational Physics}, {\bf 28}(5),
   1746--1767, (jun 2020).

\bibitem{yin2018diversity}
Dong Yin, Ashwin Pananjady, Max Lam, Dimitris Papailiopoulos, Kannan
  Ramchandran, and Peter Bartlett, `Gradient diversity: a key ingredient for
  scalable distributed learning', in {\em Proceedings of the Twenty-First
  International Conference on Artificial Intelligence and Statistics}, eds.,
  Amos Storkey and Fernando Perez-Cruz, volume~84 of {\em Proceedings of
  Machine Learning Research}, pp. 1998--2007. PMLR, (09--11 Apr 2018).

\bibitem{Zhang2020EmpiricalSO}
Xiao Zhang and Dongrui Wu, `Empirical studies on the properties of linear
  regions in deep neural networks', {\em ArXiv}, {\bf abs/2001.01072}, (2020).

\end{thebibliography}

\end{document}